%% file: arxiv.tex
\documentclass[12pt]{article}

\usepackage{fullpage}
\usepackage[utf8]{inputenc}
\usepackage[T1]{fontenc}
\usepackage{amsmath,amssymb,amsfonts}
\usepackage{amsthm}
\usepackage{graphicx}
\usepackage{mathtools}
\usepackage{hyperref}
\usepackage[sort]{natbib}

\usepackage{tikz}
\usetikzlibrary{arrows.meta}
\usetikzlibrary{patterns}
\usetikzlibrary{decorations.pathreplacing}
\usetikzlibrary{positioning, arrows.meta, shapes.geometric}
\usetikzlibrary{calc}

\usepackage{fontspec}
\newfontfamily\barlow{Barlow-SemiBold}[Path=./fonts/]

\definecolor{bgfig_light}{HTML}{ECF0F1}
\definecolor{bgfig1}{HTML}{D5DBDB}

\definecolor{directive}{HTML}{bf207c}
\definecolor{processor}{HTML}{6c98aa}
\definecolor{memory}{HTML}{5c8950}

\usepackage{pifont}

\newcommand{\NN}{\mathbb{N}}

\newtheorem{theorem}{Theorem}

\newtheorem{corollary}[theorem]{Corollary}

\newcommand\Vtextvisiblespace[1][.3em]{
  \mbox{\kern.06em\vrule height.3ex}
  \vbox{\hrule width#1}
  \hbox{\vrule height.3ex}}

\usepackage{algorithm}
\usepackage[noEnd=true]{algpseudocodex}

\title{Universal computation is intrinsic to\\language model decoding}

\author{Alex Lewandowski$^{1,2}$, Marlos C. Machado$^{1,2,3}$, Dale Schuurmans$^{1,2,3}$
  \vspace{2mm}\\
  $^{1}$Department of Computing Science, University of Alberta,
  $^{2}$Amii,\\
   $^{3}$Canada CIFAR AI Chair
   \vspace{-4mm}
}
\date{}

\begin{document}

\maketitle

\begin{abstract}
  \input{components/0_abstract.tex}
\end{abstract}

\section{Introduction}
\input{components/1_introduction}

\section{Autoregression and Computation}
\input{components/2_background}

\section{Extending Autoregressive Decoding to Universality}\label{sec:auto}
\input{components/3_autoregressive_universal}

\section{Demonstrating Computational Universality With a Proof-of-Simulation}
\input{components/4_procedure}

\section{Proof-of-Simulation via a System Prompt}\label{sec:prompting}
\input{components/5_prompting}

\section{Proof-of-Simulation via an Injective Codebook}\label{sec:learning}
\input{components/6_learning}

\section{Discussion and Implications}
\input{components/7_discussion}

\section*{Acknowledgments}
Sincere thanks to Jonathan Schaeffer and J. Quinn Lee for their comments and feedback on this paper.
The research is supported in part by the Natural Sciences and Engineering Research Council of Canada (NSERC), the Canada CIFAR AI Chair Program, and the Digital Research Alliance of Canada.

\bibliographystyle{apalike}
\bibliography{references.bib}

\newpage
\appendix

\input{components/8_appendix_a_related_work.tex}

\section{Extended Autoregressive Decoding as Universal Lag Systems}\label{app:bridge}
\input{components/8_appendix_b_bridge.tex}
\input{components/8_appendix_c_main}

\section{Proofs}
\setcounter{theorem}{0}  
\input{components/8_proofs_c_main.tex}

\end{document}

%% file: components/0_abstract.tex
Language models now provide an interface to express and often solve general problems in natural language, yet their ultimate computational capabilities remain a major topic of scientific debate.
Unlike a formal computer, a language model is trained to autoregressively predict successive elements in human-generated text.
We prove that chaining a language model's autoregressive output is sufficient to perform universal computation.
That is, a language model can simulate the execution of any algorithm on any input.
The challenge of eliciting desired computational behaviour can thus be reframed in terms of programmability: the ease of finding a suitable prompt.
Strikingly, we demonstrate that even randomly initialized language models are capable of universal computation before training.
This implies that training does not give rise to computational expressiveness---rather, it improves programmability, enabling a natural language interface for accessing these intrinsic capabilities.

%% file: components/1_introduction.tex
Artificial intelligence tools are increasingly used to solve challenging problems in diverse fields of science~\citep{bausch24_learn,lam23_learn,degrave22_magnet,jumper21_highl_alphaf,bellemare20_auton,mankowitz23_faster}.
Language models, in particular, are often capable of solving general problems expressed in natural language~\citep{openai23_gpt_techn_repor,team23_gemin,team24_gemin, deepseek-ai25_deeps_r1},
raising fundamental questions about the ultimate computational capabilities of these tools. 
Central to how these language models operate is autoregressive decoding: each generated output is appended to the input, producing a sequence of intermediate outputs that constitute the model's response to the initial input.
This decoding process is shaped by pre-training for one-step prediction~\citep{radford18_improv},
and by post-training to achieve desired autoregressive behaviours~\citep{ouyang22_train, zelikman22_star}.
Characterizing the computational capabilities of autoregressive decoding would thus establish what problems language models can and cannot solve---and clarify what role, if any, training plays in conferring these capabilities.
\looseness=-1

Despite their increasing capabilities, language models have been met with skepticism in the research community.
It has been argued, for example, that language models cannot plan \citep{kambhampati24_posit}, perform logical~\citep{jiang24_peek_token_bias} or mathematical~\citep{mirzadeh25_gsm_symbol} reasoning, explore in-context~\citep{krishnamurthy24_can}, or correct errors made in their own reasoning without external feedback~\citep{huang23_large_languag_model_cannot_self}.
Such claims are often supported by empirical demonstrations of systems failing to solve even trivial tasks.
Nevertheless, the fundamental question raised in light of these results is whether it is possible to formally characterize the theoretical limitations of a language model.
\looseness=-1

One important theoretical characterization is determining whether the autoregressive decoding of a language model can simulate a universal Turing machine.
This characterization is fundamental because of the Church-Turing thesis~\citep{church36,turing36_entsc}, which asserts that any form of computational behaviour can ultimately be realized by a Turing machine;
moreover, there are individual Turing machines (so-called universal Turing machines) that can exactly simulate the execution of any other Turing machine on any input.
Therefore, if one accepts that abilities such as reasoning, regardless of how they are defined, can be performed by a computer, then it follows that any universal Turing machine is capable of performing these same abilities.
\looseness=-1

In this paper, we prove that chaining the autoregressive outputs from 
existing, publicly available language models,
such as \texttt{Llama{-}4{-}17B{-}128E{-}Instruct}~\citep{llama24_llama_herd_model},
is sufficient to achieve computational universality.
Although one might surmise that such an ability arises from the exorbitant training costs, 
exceeding a hundred million dollars \citep{maslej24_artif_intel_index_repor,maslej25_artif_intel_index_repor},
 we also prove that even a \emph{randomly initialized}
 language model is capable of universal computation
through autoregressive decoding.
We establish this result by showing that
randomly initialized language models
are still able to
satisfy the same set of requirements that ensure computational universality 
for trained language models under autoregressive decoding.
\looseness=-1

Crucially, this result holds for various sequence modelling architectures: autoregressive decoding is sufficient for universal computation even before training.
Thus, we disentangle computational universality from the training process, the data, and the specific architecture.
This implies that training on natural language is primarily a mechanism for shaping how we interact with these systems, ensuring that model outputs are understandable to humans and that human instructions are understandable to them.

\begin{figure}\label{fig:paradigms}
  \centering
\includegraphics[width=0.99\linewidth]{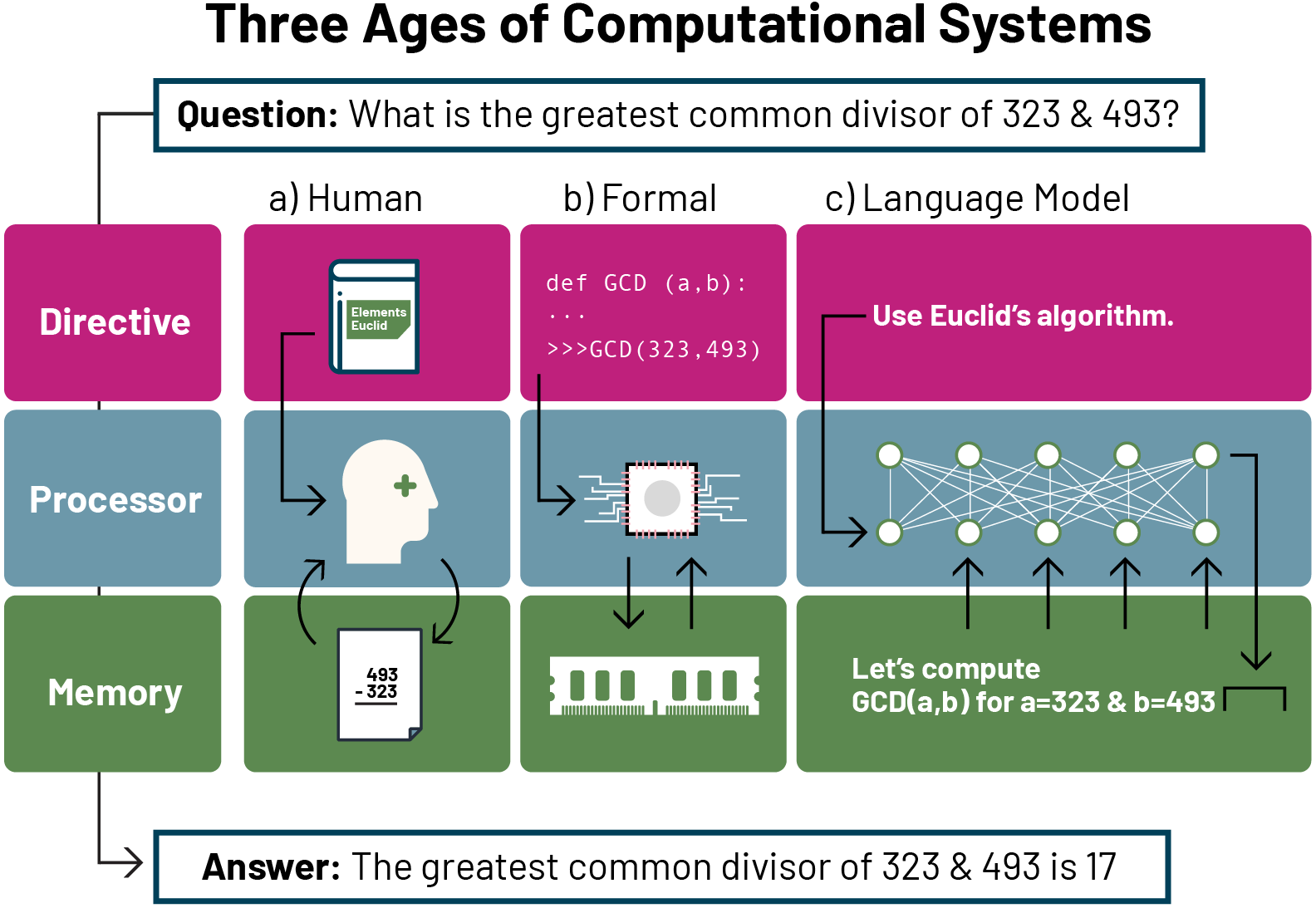}
\caption{
{\bf The three ages of computation: human, formal, and language model computation}.
  A general-purpose computational system involves three fundamental components: directive, processor, and memory.
  Given a directive, the system manipulates memory using its processor to produce an answer to a question.
  {\bf (a) Human computation}: A person uses paper as memory, manipulating it by hand according to the directive of a book.
  Before the advent of modern digital computers, humans had to manually process and manage computational operations specified by instructions written in natural language.
  {\bf (b) Formal computation}: A classical computer uses random access memory, manipulating it using a central processing unit according to the directive of a written program.
  During the second age, spanning the 80 years since the development of modern digital computers, people have had to undergo a multi-year training process to learn how to specify precise step-by-step operations for a formal machine.
  {\bf (c) Language model computation}: A language model uses a text sequence as memory, manipulating it via autoregressive decoding according to the directive of a system prompt.
  Now, in the emerging third age, language models are increasingly able to produce answers given only informal natural language specifications, without needing explicit step-by-step instructions on how to produce the answer.
}
\end{figure}

Consequently, because trained language models continue to be universal computers, the empirical failures reported in the literature cannot properly be attributed to 
inherent computational limitations of language models.
Rather, these demonstrations illustrate a failure to use natural language prompts to elicit desired, precise computations.
That is, the precision of a formal system, which leads to much-desired out-of-domain generalization, is not easily emulated in natural language.
From these findings, one can interpret language models as providing a natural language interface between humans and
computers---potentially establishing a third age in the evolution of computational systems (Figure~\ref{fig:paradigms}).
\looseness=-1

%% file: components/2_background.tex
A language model generates its response to a given input by an autoregressive decoding process, where each generated token is successively appended to the end of the input string, thus creating an extended input for the next token.
This process normally continues until a special halt-token is generated or a maximum input length  (\emph{i.e.}, context window), $N<\infty$, is reached.
To express the autoregressive process more formally, we first require a few definitions.
Consider a finite alphabet $\Sigma=\{\sigma_1,\dotsc,\sigma_K\}$,
which is assumed to contain a special halt symbol $h\in\Sigma$.
A \emph{string} is
a finite sequence of symbols, $s_1 \ldots s_n$, such that each element of the sequence belongs to the finite alphabet $\Sigma$.
Let $\Sigma^\star$ denote the set of possible strings.
Given an input string, $s_1 \ldots s_n \in \Sigma^\star$, a language model defines a conditional distribution over the next symbol, 
$p(s_{n+1}|s_1 \ldots s_n)$.
Autoregressive decoding extends the language model's one-step conditional distribution to an output string, $s_{n+1}\ldots s_{n+k}$, by the chain rule of probability:
\begin{eqnarray}\label{eq:chain rule}
p(s_{n+1}\ldots s_{n+k}|s_1\ldots s_n) & = &
p(s_{n+1}|s_1\ldots s_n)\,p(s_{n+2}|s_1\ldots s_{n+1})\cdots p(s_{n+k}|s_1\ldots s_{n+k-1})
.
\;\;\;\;
\end{eqnarray}
Therefore, an output string $s_{n+1}\ldots s_{n+k}$ can be generated from the
conditional distribution on the left-hand side
of~\eqref{eq:chain rule}
by sampling each successive
symbol from the corresponding conditional distribution on the right-hand side and appending the result to the input string,
${s_{n+1}\sim p(\cdot|s_1\ldots s_n)}$,
${s_{n+2}\sim p(\cdot|s_1\ldots s_{n+1})}$,
$\dotsc$,
${s_{n+k}\sim p(\cdot|s_1\ldots s_{n+k-1})}$.

We are interested in characterizing the computational capability of a language model under this autoregressive decoding process.
The ultimate computational capability of any realizable mechanical procedure was characterized in
the work of G\"{o}del, Turing and others \citep{goedel31_ueber_saetz_princ_mathem_system_i,church36,turing36_entsc},
summarized in the \emph{Church-Turing thesis}
that all known models of computation can be reduced to what is computable by a Turing machine \citep{benamram05}.
Turing developed his notion of a \emph{Turing machine} to settle the \emph{Entscheidungsproblem} \citep{hilbert28_grund_theor_logik}, \textit{i.e.}, the decision problem of determining whether any given mathematical statement is provable. 
The Turing machine is a formal model of computation that defines a procedure to simulate the execution of an algorithm using a finite transition function that reads and writes to an unbounded memory in the form of a tape. 
Its execution is determined by the tape's finite (non-blank) initial contents, where, at each iteration, the Turing machine follows its transition function by reading its current state and the current tape symbol, and then updating its state, modifying the current tape symbol, and moving the tape.
Crucially, there exists a universal Turing machine that can simulate any other algorithm by encoding a description of that algorithm on its tape.
A modern general-purpose computer exemplifies universal computation by the fact that it can be programmed to perform arbitrary computable tasks within its available resources.
\looseness=-1

%% file: components/3_autoregressive_universal.tex
\begin{figure}\label{fig:gen_auto}
\centering
\input{figures/fig2}
\caption{\textbf{
  Autoregressive decoding with a context window of size $N$ can be extended to read and write strings of arbitrary length.}
  \emph{Standard autoregressive decoding} uses a sliding context window to append its output to the end of its context.
  However, if the input is longer than the size of the context window, then the beginning is ignored.
\emph{Generalized autoregressive decoding} uses a sliding context window to read from the beginning of the input string and append to the end of the operational string, allowing the entire string to be eventually read regardless of the size of the context window.
\emph{Extended autoregressive decoding} introduces the option to produce more than one output token per step, enabling the generation of arbitrary-length strings.
\looseness=-1
}
\end{figure}
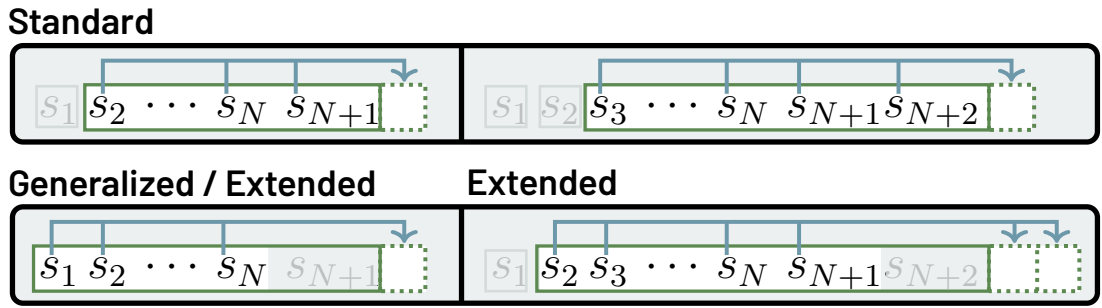

Universal computation requires the ability to manipulate memory of \emph{arbitrary} size.
Language model decoding is capable of such manipulation by simply appending generated tokens to the end of an operational string that functions as memory.
We refer to this process as \emph{generalized autoregressive decoding} to distinguish the language model's maximum context window ($N$) from the larger operational string that encapsulates the full input (possibly larger than $N$) and any additional tokens generated.
Generalized autoregressive decoding is isomorphic to conventional autoregressive decoding whenever the concatenation of the input and generated tokens does not exceed the maximum context window (\emph{i.e.}, whenever $n+k\leq N$).
We also consider \emph{extended autoregressive decoding}, where more than one token can be appended per step, which is realizable via generalized autoregressive decoding by using a special halt token $h$.
Such an application of autoregressive decoding is equivalent to the conventional decoding of a language model, which already uses a special halt token to stop its generation process,
but it allows for unbounded memory growth, which is necessary for universal computation.
These different formalization of the language model decoding process are displayed in Figure~\ref{fig:gen_auto}.
\looseness=-1

Our main results are founded on an equivalence between extended autoregressive decoding and Lag systems~\citep{wang63}, which are a variation of one of the earliest formal models of computation, Post's tag system~\cite{post43}.
Specifically, a Lag system is defined by a collection of \emph{production rules}, each of which specifies a string to string mapping.
Given an input string, the Lag system operates iteratively by observing the prefix of the input string up to a maximum length, matching this prefix to a production rule, appending the output produced by the rule to the end of the input string, and deleting some portion of the prefix from the input string.
The mechanical procedure followed by a Lag system closely resembles the operation of extended autoregressive decoding, where the language model effectively encapsulates the production rules and operation of a Lag system (we describe the exact equivalence in Appendix~\ref{app:bridge}).

We can thus understand the computational capabilities of a language model under autoregressive decoding through Lag systems.
In particular, because Lag systems can be shown to be computationally universal~\cite{wang63}, 
it is straightforward to establish that language models under extended autoregressive decoding are also provably universal.
By using a specific universal Turing machine,
$U_{15,2}$, defined on 15 states and 2 symbols \citep{neary09_four_turin}, 
it is possible to construct a corresponding universal Lag system, $L(U_{15,2})$,
that is able to exactly simulate the execution of $U_{15,2}$ on any input.
The total number of production rules in $L(U_{15,2})$ is $1857$, defined on an alphabet of $249$ symbols, where all but $14$ rules map two symbols to one symbol, with the remaining $14$ rules mapping two symbols to two symbols.
Using this universal Lag system, we can then demonstrate a language model's capability for universal computation under autoregressive decoding as outlined below.
\looseness=-1

%% file: figures/fig2.tex
\begin{tikzpicture}[scale=1.14]

\filldraw[rounded corners, line width = 2, color=black, fill=bgfig_light]
      (-175pt,50pt) rectangle +(360pt,32pt);

\fill[white]
      (-53pt,54pt) rectangle +(15pt,15pt);
\node[anchor=west] at (-180pt,90pt) {
  \scalebox{1.15}{\barlow Standard}
};
\draw[fill=white, draw=none]
      (-151pt,54pt) rectangle +(98pt,15pt);
\draw[memory, line width = 1.25]
      (-151pt,54pt) rectangle +(98pt,15pt);
\node[anchor=west] at (-170pt,60pt) {
  \scalebox{1.5}{$s_1\, s_{2}\, \cdots\, s_{N}\,\, s_{N+1} $}
};

\draw[solid, line width =2, color=black] (-26pt, 50pt) -- (-26pt, 82pt);

\draw[->, line width =1.5, color=processor] (-145pt, 66pt) -- (-145pt, 77pt) -- (-45pt, 77pt) -- (-45pt, 70pt);
\draw[line width=1.5, color=processor] (-104pt, 66pt) -- (-104pt, 77pt);
\draw[line width=1.5, color=processor] (-81pt, 66pt) -- (-81pt, 77pt);
\draw[dotted, line width = 1.5, color=memory]
      (-53pt,54pt) rectangle +(15pt,15pt);
\fill[bgfig_light, opacity=0.8]
      (-167pt,55pt) rectangle +(13pt,13pt);
\draw[bgfig1, line width = 1]
      (-167pt,55pt) rectangle +(13.5pt,13.5pt);

\fill[color=white]
      (148pt,54pt) rectangle +(15pt,15pt);
\draw[fill=white, draw=none]
      (16pt,54pt) rectangle +(134pt,15pt);
\draw[memory, line width = 1.25]
      (15pt,54pt) rectangle +(134pt,15pt);
\node[anchor=west] at (-20pt,60pt) {
  \scalebox{1.5}{$s_1\, s_{2}\, s_{3}\, \cdots\, s_{N}\,\, s_{N+1}s_{N+2} $}
};

\draw[->, line width=1.5, color=processor] (20pt, 66pt) -- (20pt, 77pt) -- (156.5pt, 77pt) -- (156.5pt, 70pt);
\draw[line width=1.5, color=processor] (62pt, 66pt) -- (62pt, 77pt);
\draw[line width=1.5, color=processor] (86pt, 66pt) -- (86pt, 77pt);
\draw[line width=1.5, color=processor] (119pt, 66pt) -- (119pt, 77pt);

\draw[dotted, line width = 1.5, color=memory]
      (149pt,54pt) rectangle +(15pt,15pt);

\fill[bgfig_light, opacity=0.8]
      (-18pt,55pt) rectangle +(13.5pt,13.5pt);
\draw[line width = 1, color=bgfig1]
      (-18pt,55pt) rectangle +(13.5pt,13.5pt);

\fill[bgfig_light, opacity=0.8]
      (0pt,55pt) rectangle +(13.5pt,13.5pt);
\draw[line width = 1, color=bgfig1]
      (0pt,55pt) rectangle +(13.5pt,13.5pt);

\begin{scope}[yshift=-0.3cm]
\node[anchor=west] at (-180pt,45pt) {
  \scalebox{1.15}{\barlow Generalized / Extended}
};

\node[anchor=west] at (-28pt,45pt) {
  \scalebox{1.15}{\barlow Extended}
};

\draw[rounded corners, line width = 2, color=black, fill=bgfig_light]
      (-175pt,5pt) rectangle +(360pt,32pt);

\fill[white]
      (-53pt,9pt) rectangle +(15pt,15pt);
\draw[fill=white, draw=none]
      (-168pt,9pt) rectangle +(78pt,15pt);
\node[anchor=west] at (-170pt,15pt) {
  \scalebox{1.5}{$s_1\, s_{2}\, \cdots\, s_{N}\,\, s_{N+1} $}
};
\fill[bgfig_light, opacity=0.8]
      (-90pt,9pt) rectangle +(36pt,15pt);
\draw[memory, line width = 1.25]
      (-168pt,9pt) rectangle +(115pt,15pt);

\draw[solid, line width =2, color=black] (-26pt, 5pt) -- (-26pt, 37pt);

\draw[->, line width=1.5, color=processor] (-162pt, 21pt) -- (-162pt, 32pt) -- (-45pt, 32pt) -- (-45pt, 25pt);
\draw[line width=1.5, color=processor] (-105pt, 21pt) -- (-105pt, 32pt);
\draw[line width=1.5, color=processor] (-145pt, 21pt) -- (-145pt, 32pt);
\draw[dotted, line width = 1.5, color=memory]
      (-53pt,9pt) rectangle +(15pt,15pt);

\fill[white]
      (0pt,9pt) rectangle +(113pt,15pt);
\node[anchor=west] at (-20pt,15pt) {
  \scalebox{1.5}{$s_1\, s_{2}\,s_{3}\, \cdots\, s_{N}\,\, s_{N+1}s_{N+2} $}
};

\fill[white]
      (149pt,9pt) rectangle +(30pt,15pt);
\draw[dotted, line width=1.5, color=memory] (165pt, 9pt) -- (165pt, 24pt);
\draw[dotted, line width = 1.5, color=memory]
      (149pt,9pt) rectangle +(30pt,15pt);
\fill[bgfig_light, opacity=0.8]
      (115pt,9pt) rectangle +(33pt,15pt);
\draw[memory, line width = 1.25]
      (-1pt,9pt) rectangle +(150pt,15pt);

\draw[->, line width=1.5, color=processor] (5pt, 21pt) -- (5pt, 32pt) -- (171.5pt, 32pt) -- (171.5pt, 25pt);

\draw[line width=1.5, color=processor] (22pt, 21pt) -- (22pt, 32pt);
\draw[line width=1.5, color=processor] (62pt, 21pt) -- (62pt, 32pt);
\draw[line width=1.5, color=processor] (86pt, 21pt) -- (86pt, 32pt);

\draw[<-, line width=1.5, color=processor] (157.5pt, 25pt) -- (157.5pt, 32pt);

\fill[bgfig_light, opacity=0.8]
      (-18pt,9pt) rectangle +(13.5pt,13.5pt);
\draw[line width = 1, color=bgfig1]
      (-18pt,9pt) rectangle +(13.5pt,13.5pt);
      \end{scope}

\end{tikzpicture}


%% file: components/4_procedure.tex
Our main results are based on
a \emph{proof-of-simulation} strategy that is able to show a given language model can exactly simulate a universal Lag system via extended autoregressive decoding, thus establishing its computational universality.
Formally, we aim to demonstrate computational universality of a candidate language model, ${M: \Phi^\star \rightarrow \Phi^\star}$, specifying a mapping between strings defined over some finite symbol set $\Phi$.
This is achieved by showing that 
extended autoregressive decoding of $M$ is able to exactly simulate the behaviour of the universal Lag system, $L(U_{15,2}): \Sigma^\star \rightarrow \Sigma^\star$, specifying a map between strings on a distinct symbol set $\Sigma$.
\looseness=-1

\begin{figure}\label{fig:proof-of-simulation}
  \centering
\includegraphics[width=0.99\linewidth]{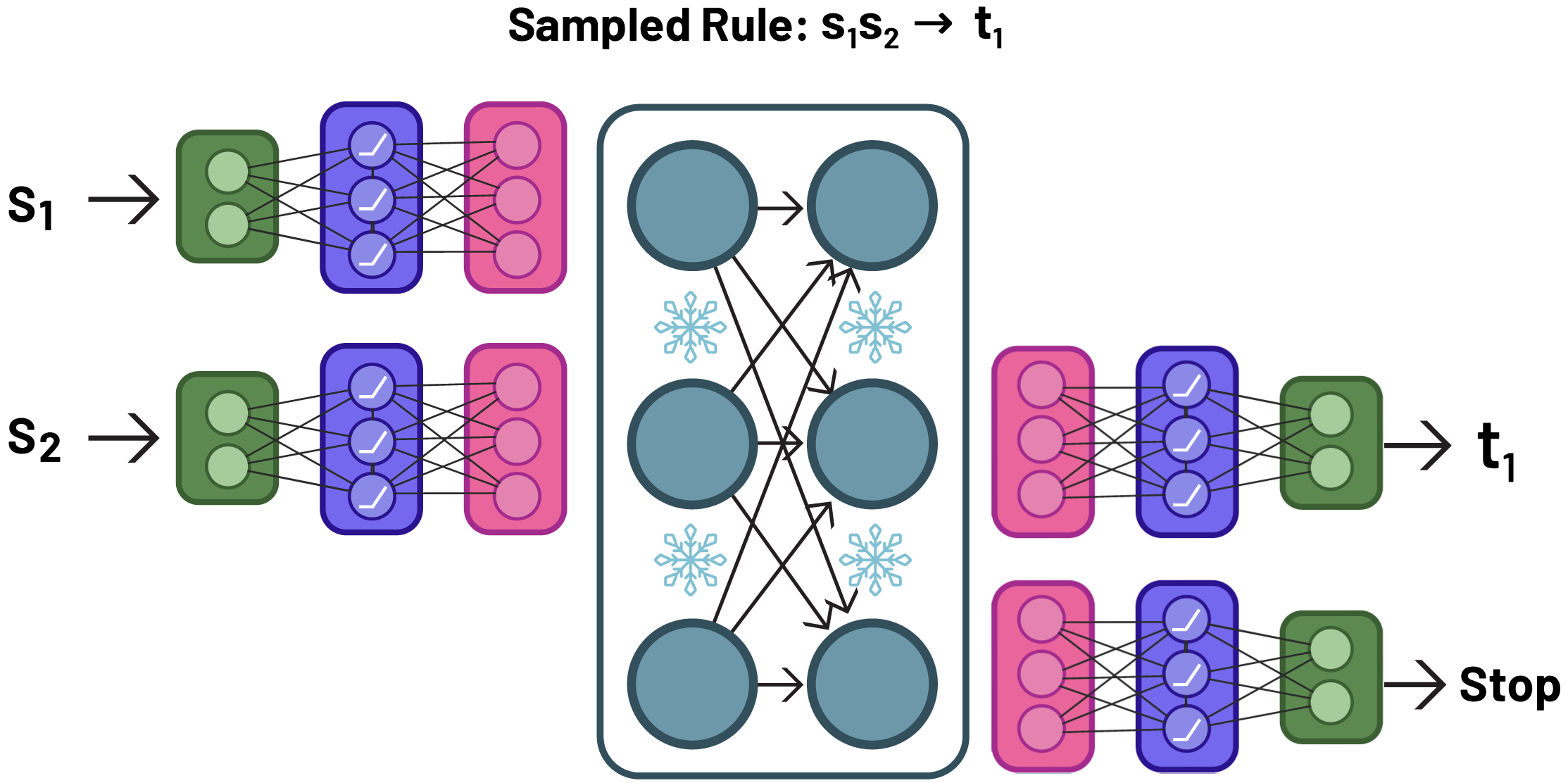}
\caption{\textbf{A language model under autoregressive decoding is computationally universal if it can simulate the universal Lag system.} Computational universality is achieved when the language model can execute each of the universal Lag system's production rules. For a one-output rule, the language model must map a two-symbol input to an output symbol $t_{1}$, and then the special halt symbol that stops generation. For a two-output rule, the language model must also produce a second output $t_{2}$ before producing the special halt symbol; this case is not shown in the figure.
}
\end{figure}

Because $M$ operates on a different symbol set than $L(U_{15,2})$, we must ensure that there exists a correspondence between the symbols in $\Sigma$ and the strings
in $\Phi^*$ that preserves the identity of the distinct symbols in $\Sigma$.
Such a correspondence can be realized by an \emph{injective} mapping $E: \Sigma \rightarrow \Phi^*$
such that there exists a reverse mapping $D:\Phi^*\rightarrow\Sigma$ satisfying $D(E(\sigma))=\sigma$ for all $\sigma\in\Sigma$.
The proof-of-simulation then proceeds by establishing that there exists a system prompt $S\in\Phi^*$ (possibly empty) such that,
for every rule of the form ${s_1s_2\rightarrow t_1}$ or ${s_1s_2\rightarrow t_1t_2}$ in $L(U_{15,2})$, the language model produces $M(SE(s_1)E(s_2))\mapsto E(t_1)E(h)$ or respectively $M(SE(s_1)E(s_2))\mapsto E(t_1)E(t_2)E(h)$ under autoregressive decoding of $M$, as depicted in Figure~\ref{fig:proof-of-simulation}.
If this can be done, then for any string $\gamma_1 \dotso \gamma_{n-1}\in\Sigma^*$, iterating this decoding of $M$ on $E(\gamma_1)\dotso E(\gamma_{n-1})$ (and dropping the halt tokens $E(h)$) exactly simulates the execution of $L(U_{15,2})$ on $\gamma_1 \dotso \gamma_{n-1}$.

We pursue two distinct strategies for establishing such a proof-of-simulation.
The first strategy (Section~\ref{sec:prompting}) is to use a nonempty system prompt, $S\in\Phi^*$, that
drives
the language model to correctly realize every rule in $L(U_{15,2})$.
Such a technique can be used to show that an existing, already trained language model is computationally universal.
The second strategy we develop (Section~\ref{sec:learning}) is to learn an injective remapping of the symbols through an encoder-decoder pair $(E,D)$, constituting a codebook, that verifies a given language model is able to correctly realize every rule in $L(U_{15,2})$. 
This second technique is used to show that \emph{randomly initialized} language models are computationally universal.
We find that the latter demonstration continues to succeed for alternative architectures.
These findings allow us to disentangle the role of the model architecture, the training procedure, and the training data from the ultimate computational  capability of a language model.

%% file: components/5_prompting.tex
We first show that autoregressive decoding of a trained language model is computationally universal by finding a system prompt that establishes a proof-of-simulation.
Specifically, we find a system prompt that drives the language model to correctly execute each of production rules from the universal Lag system.
In this case, the injective codebook $(E,D)$ is manually defined, representing each symbol from the universal Lag system as a distinct token string for the language model.
We query the language model using a standard protocol, where a fixed system prompt encodes a directive that is prepended to an active context window that slides
over the operational string.
At initialization, the operational string is set to the input and the context window is set to the start of the operational string.
On each iteration,
the next input given to the language model is constructed by concatenating the system prompt with the current context window, then
the autoregressive output from the language model (encoding one or two symbols, terminated by a special halt symbol) is appended to the end of the operational string (excluding the encoding of the halt symbol $h$), and finally the sliding context window is advanced one symbol in the operational string.
This prompting strategy emulates the procedure followed by the Lag system, and is commonly used to direct language model behaviour, such as in the recently proposed streaming approach~\cite{xiao24_effic_stream_languag_model_atten_sinks}.
Additional details can be found in Methods~\ref{methods:prompt}.

By finding a system prompt that provides a proof-of-simulation, we are able to prove that autoregressive decoding of a trained language model, such as {\tt Llama{-}4{-}17B{-}128E{-}Instruct}, is a general-purpose computer under the Church-Turing thesis.
We find the specific representation of the system prompt for this language model by an automated trial-and-error search process that combines a natural language description of the task with feedback based on the rules that the model had previously failed to correctly execute.
To verify the language model's behaviour, we fix all random seeds and greedily sample from the output distribution using a temperature of zero, which ensures that the language model output is both deterministic and discrete.
Under these settings, we have that the final system prompt discovered for {\tt Llama{-}4{-}17B{-}128E{-}Instruct} is able to drive the trained language model to correctly execute each of the universal Lag system's $1857$ production rules,
a result that can be independently verified.%
\footnote{The open-weight language model {\tt Llama{-}4{-}17B{-}128E{-}Instruct} can be accessed via the Hugging Face API\@.
 All relevant code, including the system prompt and verification script for {\tt Llama{-}17B{-}128E{-}Instruct} will be made publicly available to enable independent verification of the claims made in this paper.}
Notably, given the system prompt, this result requires only extended autoregressive decoding with the standard query protocol, without any external memory beyond the operational string being processed or any additional training.

%% file: components/6_learning.tex
Next we demonstrate how autoregressive decoding of a \emph{randomly initialized} language model can be proved to be computationally universal by learning an injective codebook that establishes the proof-of-simulation.
In particular, we recover such a codebook by jointly training encoder and decoder networks to both (i) establish an injective function from the symbols of the universal Lag system to input symbols for a randomly initialized language model, and (ii) elicit correct responses for each one of the $1857$ contexts for rules in $L(U_{15,2})$.
In this case we do not augment the language model's input with a prepended system prompt.
Similar to the verification of a trained language model's behaviour, we ensure that the randomly initialized output is deterministic and discrete by greedily sampling the output response.
Crucially, throughout this process, the language model remains randomly initialized and frozen---completely untrained.
We only train the encoder and decoder networks to learn an injective symbol remapping that drives the randomly initialized language model to correctly execute each of the universal Lag system's $1857$ production rules.

\begin{figure}\label{fig:untrained}
  \centering
\includegraphics[width=0.32\linewidth]{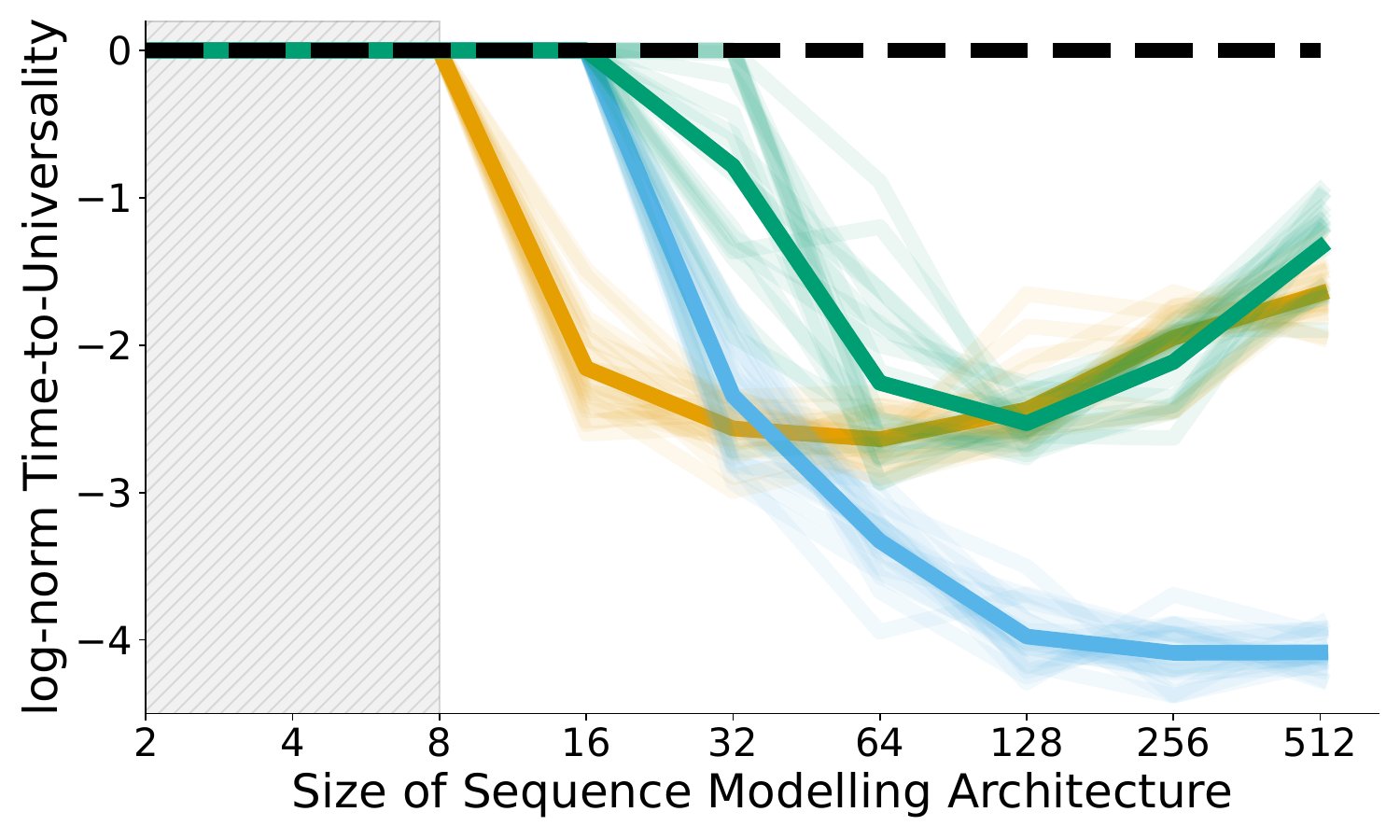}
\includegraphics[width=0.32\linewidth]{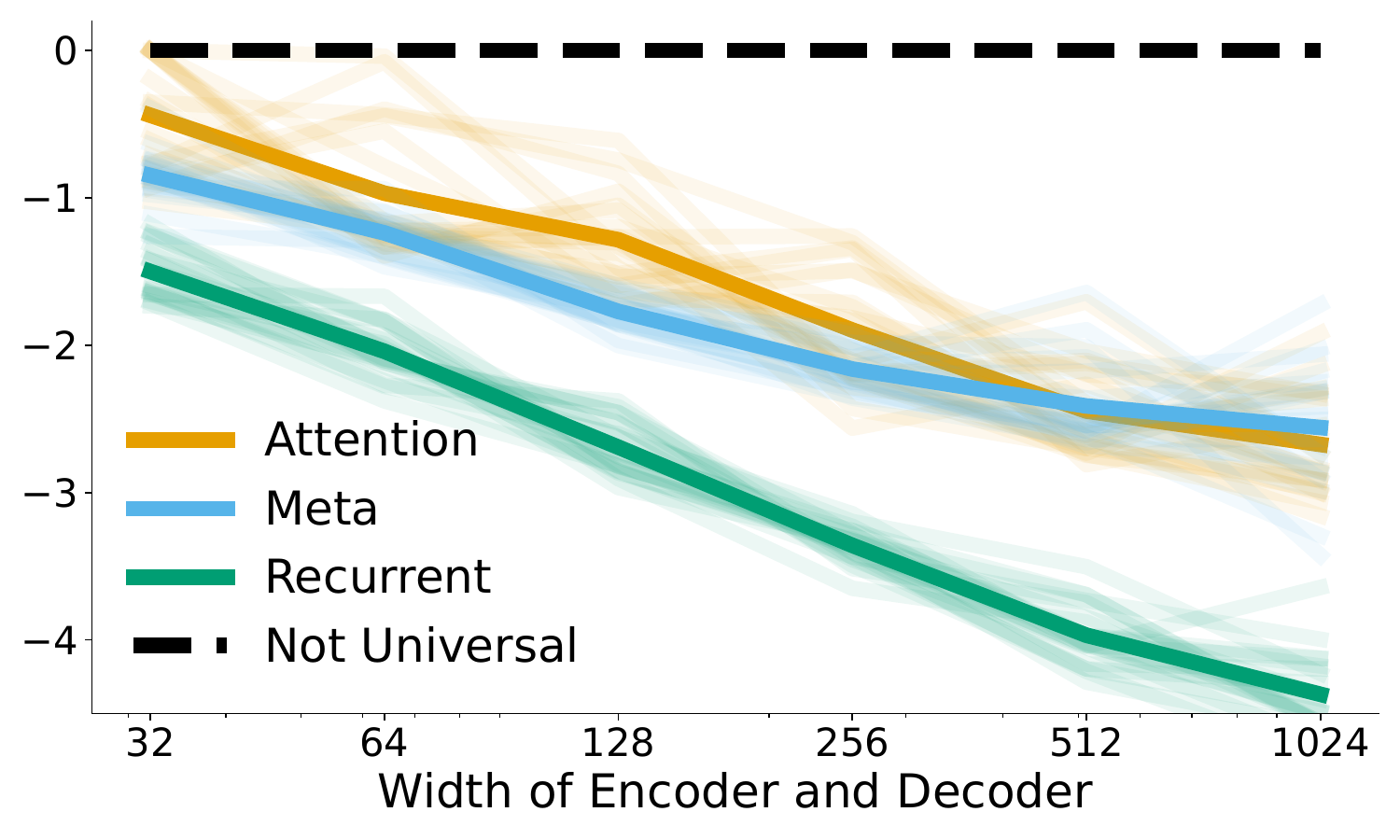}
\includegraphics[width=0.32\linewidth]{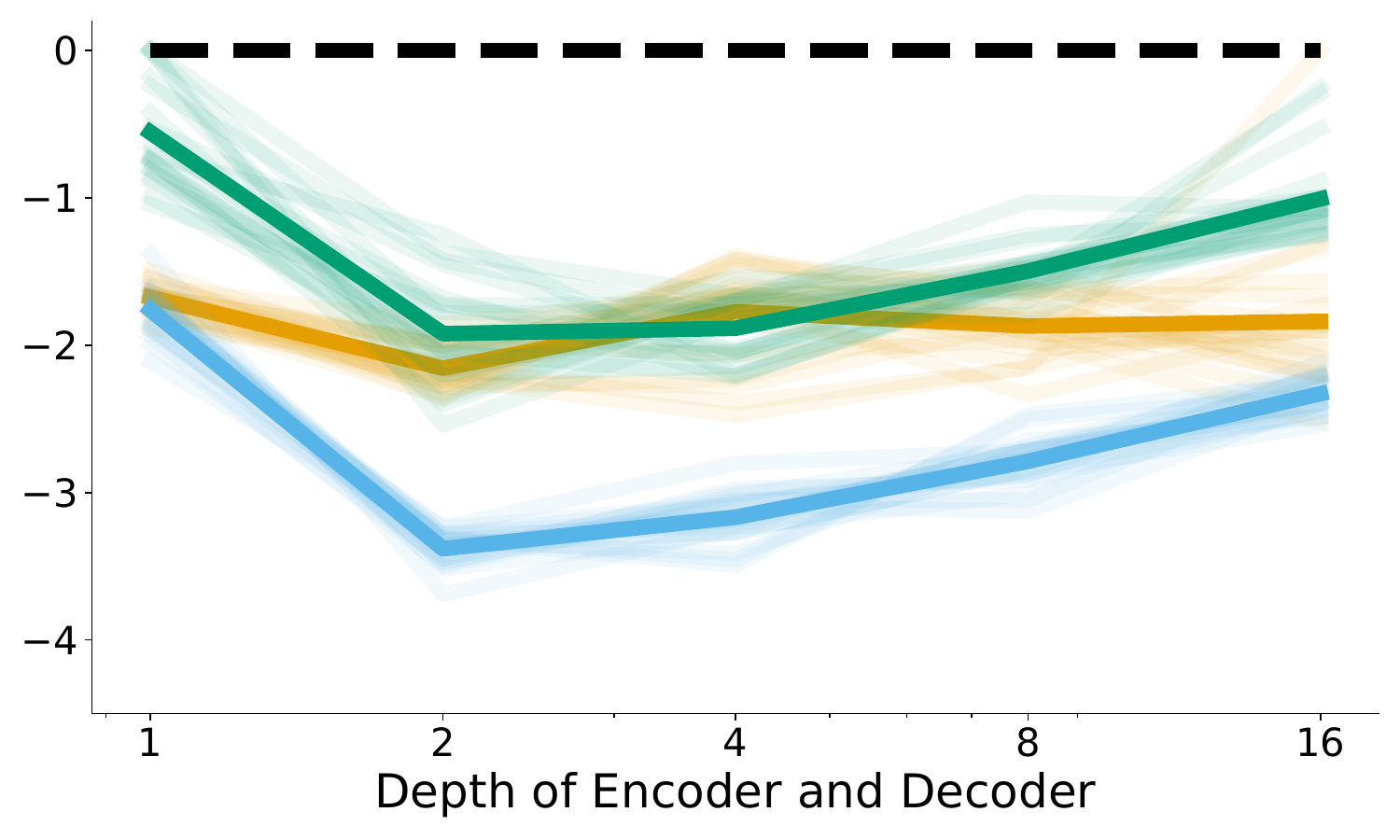}
\caption{\textbf{Randomly initialized language models achieve universal computation above an architecture-specific minimal size.}
We measure log time-to-universality: the log-normalized number of training iterations to learn a codebook that drives a randomly initialized language model to correctly execute all production rules.
Lower values indicate faster convergence, and a value of $0$ indicates that the iteration budget was reached without finding a valid codebook. 
 The faded lines show individual runs; dark lines show the average.
}
\end{figure}

The results in Figure~\ref{fig:untrained} demonstrate, by proof-of-simulation, that various randomly initialized language models are computationally universal under extended autoregressive decoding.
These findings cover several commonly used sequence modelling architectures, such as multi-head self-attention models \citep{vaswani17_atten_is_all_you_need} and recurrent networks.
We also consider an online meta-gradient method \citep{finn17_model, veeriah17_cross}, which, while less commonly used, has been shown to match the expressiveness of recurrent networks in a specific meta-learning setting~\citep{finn18_meta_learn_univer}.
For each architecture, we found that once an architecture-specific minimal size has been surpassed, various encoder and decoder networks are able to recover an injective codebook that establishes a proof-of-simulation (Figure~\ref{fig:untrained}, left). Thus, we have verified the computational universality of these models, at their initialization, under extended autoregressive decoding.

These proof-of-simulation demonstrations with randomly initialized language models show that universal computation does not require learning from massive amounts of data, nor is it tied to any specific sequence modelling architecture.
That is, for a variety of sequence modelling architectures, there exists an injective codebook that drives a randomly initialized language model to simulate a universal Lag system.
Such findings show that extended autoregressive decoding is broadly sufficient for universal computation, when applied to typical sequence modelling architectures.

%% file: components/7_discussion.tex
Our findings demonstrate that computational universality is an intrinsic property of autoregressive decoding, implying that training on human-generated text provides access to these capabilities through natural language.
That is, autoregressive decoding with a language model provides a natural interface to computation by treating its operational string as memory, which it manipulates by appending its output.
To achieve universal computation, which in principle requires unbounded memory, we have shown how autoregressive decoding can be simply chained to extend this operational string.
We proved that this form of universal computation can be simulated by a large, trained language model using a system prompt of natural language instruction.
Remarkably, we have also found that even small, untrained language models are capable of universal computation through autoregressive decoding.
This result thus disentangles computational universality from training on natural language data.
While both trained and untrained language models exhibit universal computational ability, they differ in how this is accessed: through natural language for a trained language model, or through a constructed injective codebook for a randomly initialized language model.
\looseness=-1

Characterizing a model architecture's computational capability has been a longstanding pursuit in artificial intelligence, often with the goal of establishing that a particular model has the capabilities of a universal Turing machine.
These pursuits are motivated by the possibility of artificial intelligence systems that learn formal algorithms to achieve more reliable problem-solving.
However, previous results that demonstrate computational universality have either relied on artificial assumptions such as infinite precision weights \citep{siegelmann92_comput_power_of_neural_nets}, or external mechanisms to manage memory \citep{chung21_turin_compl_bound_precis_recur_neural_networ, hobbs15_implem}.
Other results have characterized more realistic model architectures by addressing questions of representability~\citep{weiss18_rnns,merrill20_rnn} or learnability~\citep{wei22_statis} within a class of possible model instantiations, without asserting the capabilities of any particular model.
We provide a detailed discussion of all these related works in the supplementary materials.
The results in this paper, by contrast, do not rely on any assumptions beyond the actual instantiation of a given language model, and prove that existing, publicly available language models already exhibit universal computational ability.
\looseness=-1

In practice, using a language model as a computer involves searching for prompts that produce desired computational behaviour.
The name for this search process, prompt engineering~\citep{chen25_unleas}, reflects both the prompt's generality in encoding directives for behaviour~\citep{brown20_languag} and the heuristics needed to represent such directives~\citep{kojima22_large}.
Computational universality ensures the existence of a prompt for any computable behaviour, implying that failure to elicit a particular behaviour from a language model can, in principle, be attributed to a failure of prompt engineering.
However, manually engineering a prompt for a given directive can be challenging when the requisite prompt is a long sequence with little structure.
Even automated approaches to prompt engineering have been limited by the difficulty in searching over long sequences~\citep{li21_prefix_tunin}, the need for labelled data to learn prompts~\citep{shin20_autop}, and challenges in developing effective heuristics for generating prompt candidates~\citep{zhou23_large_languag_model_human_level_promp_engin}.

Given the potential difficulty of prompt engineering, programmability (the degree of ease in directing behaviour) is an important criterion for evaluating an interface to computation.
As we have shown in achieving universality with a system prompt for trained language models, searching for a prompt that produces a target behaviour can be feasible.
That is, for a user who understands natural language, a trained language model is more programmable than an untrained one.
Programmability also reframes the effectiveness of prompting with
input-output demonstrations~\citep{liu22_what_makes_good_in_contex},
chain-of-thought demonstrations~\citep{wei22_chain},
tokens for thinking~\citep{goyal24_think},
or phrases that encourage thinking step-by-step~\citep{kojima22_large,nye22_show_your_work}.
Language models prompted with these methods better interpret their computational directive and are more likely to produce the desired behaviour; they are more programmable.

With the understanding of language models as universal computers, one can improve their effectiveness by improving their programmability.
Further training of pre-trained language models can be viewed as implicitly improving their programmability by shifting the responsibility of prompt engineering away from the user.
For example, both supervised fine-tuning~\citep{wei22_finet_languag_model_zero_shot_learn,iv22_cuttin_down_promp_param,chung24_scalin} and reinforcement learning from human feedback~\citep{christiano17_deep,stiennon20_learn,ouyang22_train,bai22_const_ai,bai22_train_helpf_harml_assis_reinf} help guide autoregressive decoding toward outputs that would otherwise require well-engineered prompts. Language models can also be post-trained to produce chain-of-thought behaviour that explores the space of prompts in a process that mimics reasoning~\citep{zelikman22_star}.
These so-called reasoning models use reinforcement from verifiable rewards to autonomously search for better prompts~\citep{deepseek-ai25_deeps_r1,openai24_openai_system_card}, thus improving their programmability.
\looseness=-1

The programmability of language models has a deep historical root in Leibniz's vision of mechanizing thought.
Before the formal development of computing theory, Leibniz envisioned a systematic unification of knowledge through a universal science (\emph{scientia universalis}) comprising a universal language (\emph{characteristica universalis}) and a mechanical reasoning procedure (\emph{calculus ratiocinator})~\citep{leibniz66_disser_arte_combin,mugnai18_ars_charac}.
This vision anticipated what we now interpret as the ultimate programmable system, one that could resolve questions by translating them into a universal language and then executing a mechanical reasoning procedure.
Our results suggest that language models provide a way to approximate this ideal with natural language.
Computational universality ensures that any computable reasoning procedure is realizable through autoregressive decoding,
whereas training on natural language improves programmability so that such procedures are easier to access.
Though undecidability precludes Leibniz's ultimate vision,
and computational complexity constrains what behaviours we can practically elicit,
increasingly programmable language models are establishing a third age of computational systems---one where natural language itself becomes the interface between human thought and universal computation.
\looseness=-1

%% file: components/8_appendix_a_related_work.tex
\section{Additional Related Work}\label{app:related}
Among previous theoretical characterizations of the computational capabilities of model architectures, McCulloch and Pitt's artificial neuron is one of the first significant achievements \citep{mcculloch43}.
Networks of such artificial neurons were shown capable of computing any boolean function, and were later adapted to learning in multi-layer perceptrons \citep{hebb49,rosenblatt58}.
Recurrent architectures that generalize multi-layer perceptrons to sequences were subsequently shown capable of expressing universal computation, but these prior results depend on infinite precision weights \citep{siegelmann92_comput_power_of_neural_nets}, or external memory \citep{chung21_turin_compl_bound_precis_recur_neural_networ, hobbs15_implem}.
Real recurrent networks are computationally limited in practice, with finite precision weights that can restrict computational capability to that of a finite-state machine~\citep{korsky19_comput_power_rnns}, thus limiting what can be represented and learned \citep{weiss18_rnns,merrill20_rnn}.
It remains unknown whether universality is achieved by the conventional training and operation of a recurrent network, which has motivated several previous attempts to re-architect such networks to directly mimic the associative memory manipulations of a formal computer \citep{graves16_hybrid,grefenstette2015learning,kurach16_neural,kaiser16_neural_gpus}.

Related theoretical results have been demonstrated for transformer architectures approximating those used in current language models, establishing computational universality with infinite precision weights~\citep{perez19_turin_compl_moder_neural_networ_archit,bhattamishra20_comput_power,perez21_atten}, characterizing a programming language for the functions that are learnable by a transformer~\citep{weiss21_think_like_trans}, achieving programmable computation using hardcoded weights with looped transformer execution~\citep{giannou23_looped}, developing a finer-grained characterization of circuit complexity~\citep{merrill22_satur}, extending the execution of such circuits to chain-of-thought sequences~\citep{feng23_towar,merrill24_expres_power_trans_chain_thoug,li24_chain_thoug_empow_trans_solve}, considering statistical approximations to universal Turing machines~\citep{wei22_statis,malach24_auto_regres}, and finally improving generalization by training a transformer to mimic the execution of a Turing machine \citep{hou25_univer_lengt_gener_turin_progr}.
The computational capabilities of other architectural modifications have also been considered, such as linear approximations to transformers~\citep{irie23_pract_comput_power_linear_trans}, state-space models that have a recurrent execution mode~\citep{merrill24}, and universal transformers which explicitly combine self-attention with recurrent connections in a single architecture~\citep{dehghani19_univer_trans}.
These latter modifications inherit similar computational capabilities and limitations, but can improve representational capacity in practical settings \citep{bhattamishra2024separations}.

More recently, it was shown that a conventionally trained and operated language model can achieve universal computation when augmented with an external associative memory~\citep{schuurmans23_memor_augmen_large_languag_model_comput_univer}.
By contrast, in this paper, we establish that existing language models, such as the publicly available \texttt{Llama{-}4{-}17B{-}128E{-}Instruct} and, perhaps more surprisingly, randomly initialized language models, provably exhibit universal computational ability through autoregressive decoding, without any further assumptions or modifications.

%% file: components/8_appendix_b_bridge.tex
The technical correspondence between autoregressive language models and lag systems was established in our previous work~\citep{schuurmans24_autor_large_languag_model_comput_univer}.
In particular, that work demonstrates that generalized autoregressive decoding corresponds to the execution of a linear bounded automaton, and that extended autoregressive decoding corresponds to the execution of a Turing machine.
While the universality of Lag systems has been known since its introduction by~\cite{wang63}, our previous work involves a novel proof of universality which shows how lag systems can simulate a circular queue and how such a circular queue can facilitate bidirectional control over memory access.
Consequently, the technical material provides an explicit construction of a compiler for translating a particular Turing machine into a set of production rules for the lag system.
Using this compiler, it is possible to translate a universal Turing machine, such as $U_{15,2}$ \citep{neary09_four_turin} into a corresponding universal Lag system $L(U_{15,2})$.
The results in the present paper use this technical correspondence, along with the explicit construction of the universal Lag system, to establish a proof-of-simulation procedure for demonstrating the computational capabilities of a language model.

%% file: components/8_appendix_c_main.tex
\section{The Proof-of-Simulation Procedure}\label{appendix:procedure}
The results below establish computational universality of extended autoregressive decoding of various language models by proving exact simulation of the universal Lag system $L(U_{15,2})$.
The overall proof structure is based on a simulation strategy where each successive prompt is comprised of two components:
a system prompt $S\in\Phi^*$, 
and a ``sliding window context''
where the encoding of the next symbol pair 
from the input sequence is appended.
This prompt structure
is reminiscent of the recent streaming model proposed by~\cite{xiao24_effic_stream_languag_model_atten_sinks}.
On each iteration,
the encoding of the next symbol pair 
from the sequence is appended to the system
prompt and given to the language model as input;
the output of the language model 
is then appended to the end of the
operational string (dropping the encoding of the halt token).
This suffices to achieve an exact simulation of $L(U_{15,2})$.

\begin{theorem}\label{thm:proof-of-simulation}
For a given language model $M:\Phi^*\rightarrow\Phi^*$, 
assume there exists an injective mapping $E:\Sigma\rightarrow\Phi^*$ such that there exists a reverse mapping $D:\Phi^*\rightarrow\Sigma$ satisfying $D(E(\sigma))=\sigma$ for all $\sigma\in\Sigma$.
Assume furthermore there exists a system prompt $S\in\Phi^*$ (possibly empty) such that for every rule of the form $s_1s_2\rightarrow t_1$ (respectively $s_1s_2\rightarrow t_1t_2$) in $L(U_{15,2})$ we obtain $M(SE(s_1)E(s_2))\mapsto E(t_1)E(h)$ (respectively $M(SE(s_1)E(s_2))\mapsto E(t_1)E(t_2)E(h)$) under extended autoregressive decoding of $M$.
Then for any string $\gamma_1\ldots \gamma_{n-1}\in\Sigma^*$,
iterating the decoding of $M$ on $E(\gamma_1)\ldots E(\gamma_{n-1})$ (dropping the halt tokens $E(h)$) exactly simulates the execution of $L(U_{15,2})$ on $\gamma_1\ldots \gamma_{n-1}\#$.
\end{theorem}

Note that in these proofs, to ensure a deterministic simulation, we use a decoding temperature of zero and fix all the random seeds that define the language model's behaviour.

\section{Designing a System Prompt for Simulating a Universal Lag System}\label{methods:prompt}

Our first result demonstrates that existing trained language models can
simulate the execution of the universal Lag system $L(U_{15,2})$
on any input string.
This will be achieved by developing a specific system prompt $S\in\Phi^*$ 
that provides the full rule set
(expressed over the token pair encoding)
to the language model,
which we will then show drives extended autoregressive (greedy) decoding to exactly
satisfy the proof-of-simulation Theorem~\ref{thm:proof-of-simulation}.

To construct an injective mapping in this setting we
first note that, to implement extended autoregressive decoding, we need 
an implicit latent halt token $h$ outside the base alphabet of
$249$ symbols.
Second, 
note that the $249$ symbols used by
$L(U_{15,2})$ are not in a convenient format for this purpose,
as each symbol is a triple that requires auxiliary characters to describe
(such as parentheses and commas).
Rather than have the language model
generate extraneous syntactic detail,
we streamline the task by introducing a simple invertible mapping $E_l:\Sigma\cup\{h\}\rightarrow\Phi^2$
between the $250$ triples in $\Sigma\cup\{h\}$ and $250$ token pairs in $\Phi^2$.
In particular, we leverage a simple bijection
where each triple is assigned a unique token pair,
implementable by two dictionaries:
the encoding dictionary $E_l:\Sigma\rightarrow\Phi^2$ that maps triples to token pairs,
and an inverse decoding dictionary $D_l:\Phi^2\rightarrow\Sigma$ that maps token pairs back to triples,
ensuring that $E_l(D(_l\sigma))=\sigma$ for all $\sigma\in\Sigma$.

The proof-of-simulation then proceeds 
as outlined in Theorem~\ref{thm:proof-of-simulation},
encoding the Turing machine computation as a string of token pairs,
rather than a string of triples.

Previously, we established the proof-of-simulation for {\tt Gemini{-}1.5{-}Pro{-}002} by developing a specific system prompt $S_{g}$ that provides the entire rule set from $L(U_{15,2})$ and drives the model to correctly respond to each one of the $1857$ contexts.
However,
this model has since been deprecated
and the result can no longer be independently verified by third parties.
Therefore, we have reestablished the result for an open weight model,
{\tt Llama{-}4{-}17B{-}128E{-}Instruct}, that has been released on the Hugging Face API, 
ensuring that the following result will remain independently verifiable. 
For this particular model, we developed an injective codebook $(E_l,D_l)$ and a similar system prompt $S_l$ that provides the entire rule set from $L(U_{15,2})$ to {\tt Llama{-}4{-}17B{-}128E{-}Instruct} and drives it to correctly respond to each of the $1857$ contexts.%
\footnote{
All code and data supporting the claims in this paper will be made publicly available.
}

\begin{corollary}
For {\tt Llama{-}4{-}17B{-}128E{-}Instruct},
the system prompt $S_l$ and the injective codebook $(E_l,D_l)$
satisfy Theorem~\ref{thm:proof-of-simulation}.
\end{corollary}

From this theorem, by the Church-Turing thesis, we conclude that
{\tt Llama{-}4{-}17B{-}128E{-}Instruct}
under extended autoregressive (greedy) decoding
is a general purpose computer.
Notably, no additional computational mechanisms
beyond extended autoregressive decoding are required to achieve this result.

\section{Learning an Injective Codebook for Simulating a Universal Lag System}

Next, we show that randomly initialized language models can simulate the execution of the universal Lag system on any input string.
To do so, for a given untrained language model $M:\Phi^*\rightarrow\Phi^*$
we discover an injective mapping that establishes a proof-of-simulation for $M$.
The injective mapping consists of two networks: an encoder $E:\Sigma\rightarrow\Phi$ that learns a representation for each symbol, and a decoder $D:\Phi\rightarrow\Sigma$ that learns a mapping from the encoder's representation back to the original symbol,
which together satisfy $D(E(\sigma))=\sigma$ for all $\sigma\in\Sigma$.
In this case, the discrete representation learned by the encoder is provided as input to the randomly initialized language model to produce an output, which is then decoded to a corresponding symbol.
If the encoder and decoder drive the randomly initialized language model to correctly execute each of the universal Lag system's production rules, then the randomly initialized language model can simulate the execution of the universal Lag system on the corresponding representation of its inputs.

\begin{corollary}\label{cor:randomuniv}
For a given language model $M$, if there exists a discrete symbol mapping $E:\Sigma\rightarrow\Phi$ such that for every rule $s_1s_2\rightarrow t_1$ (respectively $s_1s_2\rightarrow t_1t_2$) in $L(U_{15,2})$ we obtain $M(E(s_1)E(s_2))\mapsto E(t_1)E(h)$ (respectively $M(E(s_1)E(s_2))\mapsto E(t_1)E(t_2)E(h)$) under extended autoregressive decoding of $M$,
then for any string $\gamma_1\ldots \gamma_{n-1}\in\Sigma^*$, iterating the extended autoregressive decoding of $M$ on $E(\gamma_1)\ldots E(\gamma_{n-1})$ (dropping the halt tokens $E(h)$) will exactly simulate the execution of $L(U_{15,2})$ on $\gamma_1\ldots \gamma_{n-1}\#$.
\end{corollary}

We conducted several distinct verifications 
to disentangle the role of the language model architecture in establishing proofs-of-simulation via Corollary~\ref{cor:randomuniv}. 
To ensure that the output is discrete and deterministic, the output vector of the sequence modelling architecture is sampled greedily.
We ensure that this operation allows for gradient-based optimization in an approach that is similar to vector quantization, where the discrete output is chosen by rounding the output vector to the nearest element of the codebook which minimizes the L2 distance.

In particular, we considered
three different language modelling architectures for the verifications: attention networks (which use multi-head self-attention), recurrent neural networks, and meta networks (which use online meta-gradient descent to update a feed-forward neural network).
The network instantiations used in each of these architectures can vary in size with a single hyperparameter that controls the input dimension, output dimension and the size of any intermediate variables (all of which are equal).
Specifically, the attention network uses 8 heads (as is common in language models) and the meta network uses RMSProp as its inner optimizer.
Each of these sequence modelling architectures process inputs corresponding to symbols in a representation space, and produce outputs consisting of symbols in the same representation space.
However, the specific mechanism used to produce an output differs depending on the architecture.
The attention network uses the input sequence to construct a key, query and value using its randomly initialized parameters, which is then used to produce an output without any explicit state outside of the input sequence.
The recurrent network maintains an explicit state, which is updated sequentially on each input using its randomly initialized parameters.
The meta network uses the parameters of a feed-forward network (we use a linear network) as an explicit state, and updates these randomly initialized parameters through a gradient of a denoising objective.\footnote{The choice of objective for the meta network is arbitrary, and we chose a denoising objective so that the update is solely a function of the input, as is the case with the recurrent network.}
The updates to the linear network only occur along the sequence, and, just as the the initial state of a recurrent network is reset, the parameters of the linear network are reset at the beginning of each new input sequence.

To establish a proof-of-simulation for each language model, we recovered an injective mapping that satisfies Corollary~\ref{cor:randomuniv} by training
encoder and decoder neural networks using a typical feed-forward network architectures, augmented with skip connections \citep{he16_deep_resid_learn_image_recog} and layer normalization \citep{ba16_layer_normal}.
To train these encoder and decoder networks to correctly satisfy each one of the $1857$ constraints in Corollary~\ref{cor:randomuniv}, we used the Adam optimizer \citep{kingma14_adam}, with a step size of $0.0001$ where every update used the entire dataset of production rules to calculate the gradient.

All runs were conducted with a maximum iteration count of $20,000$, although we found that that proof-of-simulation could be achieved as early as iteration $1000$.
We verified the proof-of-simulation property
for 30 different random initializations of the three language modelling architectures.
While some of the architectures can achieve universality at a lower dimensionality, we found that every run for each architecture resulted in a proof-of-simulation at a dimensionality of $64$.
Figure~\ref{fig:untrained} reports log normalized time-to-universality, which is measured as the log normalized number of iterations for which all of the production rules are correctly executed.

%% file: components/8_proofs_c_main.tex
\begin{theorem}\label{thm:proof-of-simulation}
For a given language model $M:\Phi^*\rightarrow\Phi^*$, 
assume there exists an injective mapping $E:\Sigma\rightarrow\Phi^*$ such that there exists a reverse mapping $D:\Phi^*\rightarrow\Sigma$ satisfying $D(E(\sigma))=\sigma$ for all $\sigma\in\Sigma$.
Assume furthermore there exists a system prompt $S\in\Phi^*$ (possibly empty) such that for every rule of the form $s_1s_2\rightarrow t_1$ (respectively $s_1s_2\rightarrow t_1t_2$) in $L(U_{15,2})$ we obtain $M(SE(s_1)E(s_2))\mapsto E(t_1)E(h)$ (respectively $M(SE(s_1)E(s_2))\mapsto E(t_1)E(t_2)E(h)$) under extended autoregressive decoding of $M$.
Then for any string $\gamma_1\ldots \gamma_{n-1}\in\Sigma^*$,
iterating the decoding of $M$ on $E(\gamma_1)\ldots E(\gamma_{n-1})$ (dropping the halt tokens $E(h)$) exactly simulates the execution of $L(U_{15,2})$ on $\gamma_1\ldots \gamma_{n-1}\#$.
\end{theorem}

\begin{proof}
We adopt the shorthand notation $E(\gamma_1\ldots \gamma_{n-1}\#)=E(\gamma_1)\ldots E(\gamma_{n-1})E(\#)$.
Note that each rule in $L(U_{15,2})$ has the form $s_1s_2\rightarrow t_1$ or $s_1s_2\rightarrow t_1t_2$, such that $s_1,s_2,t_1,t_2\in\Sigma$.
Given an input string $u_0=\gamma_1\ldots \gamma_{n-1}\#\in\Sigma^*$, the Lag system $L(U_{15,2})$ will produce a sequence of strings $u_0,u_1,\ldots ,u_k,\ldots $ according to its successive updates.
We prove by induction that, under the assumption of the theorem statement, extended autoregressive decoding of the language model $M$ will produce a corresponding sequence of strings $E(u_0),E(u_1),\ldots ,E(u_k)\ldots $ such that $D(E(u_k))=u_k$ for all strings $u_k$, $k\in\NN$.

For the base case, we begin with the observation that $D(E(u_0))=u_0$, since it was assumed that $D(E(\sigma))=\sigma$ for all $\sigma\in\Sigma$.

Then for the induction hypothesis, assume $D(E(u_{k-1}))=u_{k-1}$ up to some iteration $k-1$.
We will show that this implies $D(E(u_{k}))=u_{k}$, which will establish the result.

At step $k-1$, let $u_{k-1}=\gamma_1\ldots \gamma_K$ for $\gamma_i\in\Sigma$.
For the corresponding update, the Lag system $L(U_{15,2})$ will match the prefix according to some rule $\gamma_1\gamma_2\rightarrow t_1$ (Case 1) or $\gamma_1\gamma_2\rightarrow t_1t_2$ (Case 2).

In Case 1, the next string produced by the Lag system will be $u_k=\gamma_2\ldots \gamma_k t_1$.
By the induction hypothesis, the language model $M$ will start with the operational string $E(\gamma_1)\ldots E(\gamma_K)$.
In a single extended autoregressive decoding step, given the prompt $SE(\gamma_1)E(\gamma_2)$, the language model $M$ will emit the response $E(t_1)E(h)$ by assumption.  After appending the response to the operational string (excluding $E(h)$) and advancing the context window one step, the resulting operational string will be $E(\gamma_2)\ldots E(\gamma_K)E(t_1)$, hence 
$D(E(\gamma_2)\ldots E(\gamma_K)E(t_1))
= D(E(\gamma_2))\ldots D(E(\gamma_K))D(E(t_1))
=\gamma_2\ldots \gamma_K t_1
= u_k$.

In Case 2, the next string produced by the Lag system will be $u_k=\gamma_2\ldots \gamma_k t_1t_2$.
By the induction hypothesis, the language model $M$ will start with the operational string $E(\gamma_1)\ldots E(\gamma_K)$.
In a single extended autoregressive decoding step, given the prompt $SE(\gamma_1)E(\gamma_2)$, the language model $M$ will emit the response $E(t_1)E(t_2)E(h)$ by assumption.  After appending the response to the operational string (excluding $E(h)$) and advancing the context window one step, the resulting operational string will be $E(\gamma_2)\ldots E(\gamma_K)E(t_1)E(t_2)$, hence
$D(E(\gamma_2)\ldots E(\gamma_K)E(t_1)E(t_2))
= D(E(\gamma_2))\ldots D(E(\gamma_K))D(E(t_1))D(E(t_2))
=\gamma_2\ldots \gamma_K t_1 t_2
= u_k$.
\end{proof}

\begin{corollary}
For {\tt Llama{-}4{-}17B{-}128E{-}Instruct},
the system prompt $S_l$ and the injective codebook $(E_l,D_l)$
satisfy Theorem~\ref{thm:proof-of-simulation}.
\end{corollary}

\begin{proof}
The proof consists of an enumeration of the $1857$ cases,
where, for each rule $s_1s_2\rightarrow y$ in $L(U_{15,2})$,
the prompt
$S_l E_l(s_1)E_l(s_2)$
is provided to the
language model and the model's response is verified to be $E_l(y)E_l(h)$.
Then, using Theorem~\ref{thm:proof-of-simulation} and the assumption that $D_l(E_l(\sigma))=\sigma$ for all $\sigma\in\Sigma$, one can conclude that
{\tt Llama{-}4{-}17B{-}128E{-}Instruct}
 is able to simulate the execution of the universal Lag system
$L(U_{15,2})$.
\end{proof}

\begin{corollary}\label{cor:randomuniv}
For a given language model $M$, if there exists a discrete symbol mapping $E:\Sigma\rightarrow\Phi$ such that for every rule $s_1s_2\rightarrow t_1$ (respectively $s_1s_2\rightarrow t_1t_2$) in $L(U_{15,2})$ we obtain $M(E(s_1)E(s_2))\mapsto E(t_1)E(h)$ (respectively $M(E(s_1)E(s_2))\mapsto E(t_1)E(t_2)E(h)$) under extended autoregressive decoding of $M$,
then for any string $\gamma_1\ldots \gamma_{n-1}\in\Sigma^*$, iterating the extended autoregressive decoding of $M$ on $E(\gamma_1)\ldots E(\gamma_{n-1})$ (dropping the halt tokens $E(h)$) will exactly simulate the execution of $L(U_{15,2})$ on $\gamma_1\ldots \gamma_{n-1}\#$.
\end{corollary}

\begin{proof}
The proof is isomorphic to the proof of Theorem~\ref{thm:proof-of-simulation}
using an empty system prompt string $S$.
\end{proof}